\pdfoutput=1
\documentclass[11pt]{article}
\usepackage[]{ACL2023}

\usepackage{times}
\usepackage{latexsym}
\usepackage[T1]{fontenc}
\usepackage[utf8]{inputenc}
\usepackage{microtype}
\usepackage{inconsolata}
\usepackage{xcolor}

\usepackage{hyperref}

\usepackage{color, colortbl}
\usepackage{booktabs}
\usepackage[algo2e,ruled,vlined]{algorithm2e} 
\usepackage{amsmath,amsfonts,bbm}
\usepackage{multirow}
\usepackage{booktabs,adjustbox}
\usepackage{svg} 
\definecolor{prompt}{RGB}{255,228,196}
\usepackage[symbol]{footmisc}

\title{Enhancing and Assessing Instruction-Following with Fine-Grained\\ Instruction Variants}

\author{Jiuding Yang $^1$\ 
\stepcounter{footnote}Weidong Guo\thanks{\ \ Corresponding author.} $^2$
\ Kaitong Yang $^2$\  
Xiangyang Li $^2$ \\
\textbf{Yu Xu} $^2$\ 
\textbf{Di Niu} $^1$\  \\
\text{$^1$University of Alberta}\\
\text{$^2$Platform and Content Group, Tencent}\\
\texttt{$^1$\{jiuding,dniu\}@ualberta.ca}\\
\texttt{$^2$\{weidongguo,kaitongyang,xiangyangli,henrysxu\}@tencent.com}
}

\begin{document}
\maketitle

\begin{abstract}
The effective alignment of Large Language Models (LLMs) with precise instructions is essential for their application in diverse real-world scenarios. Current methods focus on enhancing the diversity and complexity of training and evaluation samples, yet they fall short in accurately assessing LLMs' ability to follow similar instruction variants. We introduce an effective data augmentation technique DeMoRecon that decomposes complex instructions into simpler sub-components, modifies these, and reconstructs them into new variants, thereby preserves the original instruction's context and complexity while introducing variability, which is critical for training and evaluating LLMs' instruction-following precision. Based on DeMoRecon, we developed the FGIV dataset which contains fine-grained instruction variants of 1,773 seed instructions to both fine-tune and evaluate LLMs. Our findings show that LLMs fine-tuned with FGIV will gain significant performance boost on both ours and commonly used instructions-following benchmarks.
\end{abstract}

\section{Introduction}
\label{sec:intro}
The development of Large Language Models (LLMs) has significantly advanced Natural Language Processing (NLP), especially in natural language generation, due to their advanced understanding capabilities. While prompt engineering initially showed promise \citep{brown2020language, sahoo2024systematic, wang2023prompt}, real-world applications have revealed significant challenges, particularly in following complex instructions. To overcome these, researchers have developed techniques like instruction tuning \citep{instruction-tuning} and alignment tuning \citep{RLHF}, which enhance the precision and effectiveness of LLMs in practical scenarios.

Instruction tuning, or supervised fine-tuning (SFT), trains LLMs using high-quality instruction-response pairs, demonstrating significant benefits even with a limited dataset, as shown by the study of \citet{LIMA}, which effectively fine-tuned models with just 1,000 pairs. Another advancement, the \textsc{Evol-Instruct} technique by \citet{wizardLM}, iteratively refines instructions through interactions with ChatGPT, leading to measurable improvements across various benchmarks.

On the other hand, alignment tuning aims to improve how models discern effective from ineffective responses, aligning their outputs more closely with human preferences. Techniques like Direct Preference Optimization \cite{DPO} and Prospect Theoretic Optimization \cite{KTO} have been crucial, with applications such as the integration of GPT-4 with DPO in the work by \citet{conifer}, setting new standards in instruction-following tasks.

Despite these advancements, there remains a notable gap: the current method do not adequately capture the subtleties within instructions that only vary slightly, which can be crucial in complex application scenarios.
This gap is also evident in current evaluation benchmarks. Most benchmarks, such as IFEval \citep{IFEval}, FollowBench \citep{FollowBench}, and InfoBench \citep{InFoBench}, focus on rule-based, progressively challenging instructions, and reliability and interpretability of the evaluation on complex instructions, but fail to assess the fine nuances in instructions that are contextually similar. 

This oversight underscores a critical gap in our enhancing and assessment methods, highlighting the need for developing enhancing method and evaluation metrics that can learn and detect these finer nuances in instruction understanding. Such metrics would enable a more detailed analysis of LLMs' comprehension and execution capabilities for subtly varied instructions, offering deeper insights into model performance and facilitating more targeted refinements in instruction-based training.

To address these challenges, we propose \textbf{DeMoRecon}, an effective data augmentation method that constructs instruction variants with only small differences in the sub-instructions of the original data by \textbf{De}composing each instruction into sub-instructions, \textbf{Mo}difying them, and then \textbf{Recon}structing these into new variants. This process preserves the original context and difficulty while increasing variability, thereby improving and assessing LLMs' instruction-following capabilities

Based on DeMoRecon, we have developed the \textbf{FGIV} datasets, which \textbf{F}ine-\textbf{G}rained \textbf{I}nstruction \textbf{V}ariants of 1,773 seed instructions to both enhance and evaluate these capabilities. The FGIV-A dataset uses direct responses from GPT-4 based on the augmented instructions for DPO, whereas FGIV-R incorporates responses revised from the original dataset to maintain closeness to the initial instructions. Additionally, the FGIV-Eval dataset, which includes 170 instructions and their augmentations, leverages GPT-4 to assess the accuracy of following very similar instruction variants.

We leverage both instruction tuning and alignment tuning to assess the effectiveness of the FGIV datasets on LLMs and evaluating their performance across FGIV-Eval and established benchmarks like IFEval, InfoBench, and FollowBench. The results demonstrate notable enhancements in the precision with which LLMs follow instructions. Key contributions of our work include:
\begin{itemize}
    \item {We introduce DeMoRecon, a data augmentation method that breaks down complex instructions into simpler sub-components, modifies them, and reconstructs them into new variants. This preserves the original instruction's difficulty and context while increasing variability, thereby improving LLMs’ ability to detect nuanced differences in commands. }

    \item{Based on DeMoRecon, we develop the FGIV dataset, which includes three components (FGIV-A, FGIV-R, and FGIV-Eval), targeting both enhanced training and rigorous evaluation of LLMs' instruction-following abilities. This dataset is specifically designed to test LLMs on subtle variations within instructions, providing a refined tool for assessing nuanced comprehension and execution.}
    
    \item{Through comprehensive testing, we show that LLMs trained with FGIV outperform those trained with traditional benchmarks in nuanced instruction-following tasks. Our findings, validated using several leading LLMs including GPT-4, establish FGIV as a critical benchmark for future assessments in this field. We have made the training data and source code fully open-source to support ongoing research.}
\end{itemize}

\section{Approach}
\label{sec:method}
\begin{figure*}[ht]
\centering
\includegraphics[width=0.95\textwidth]{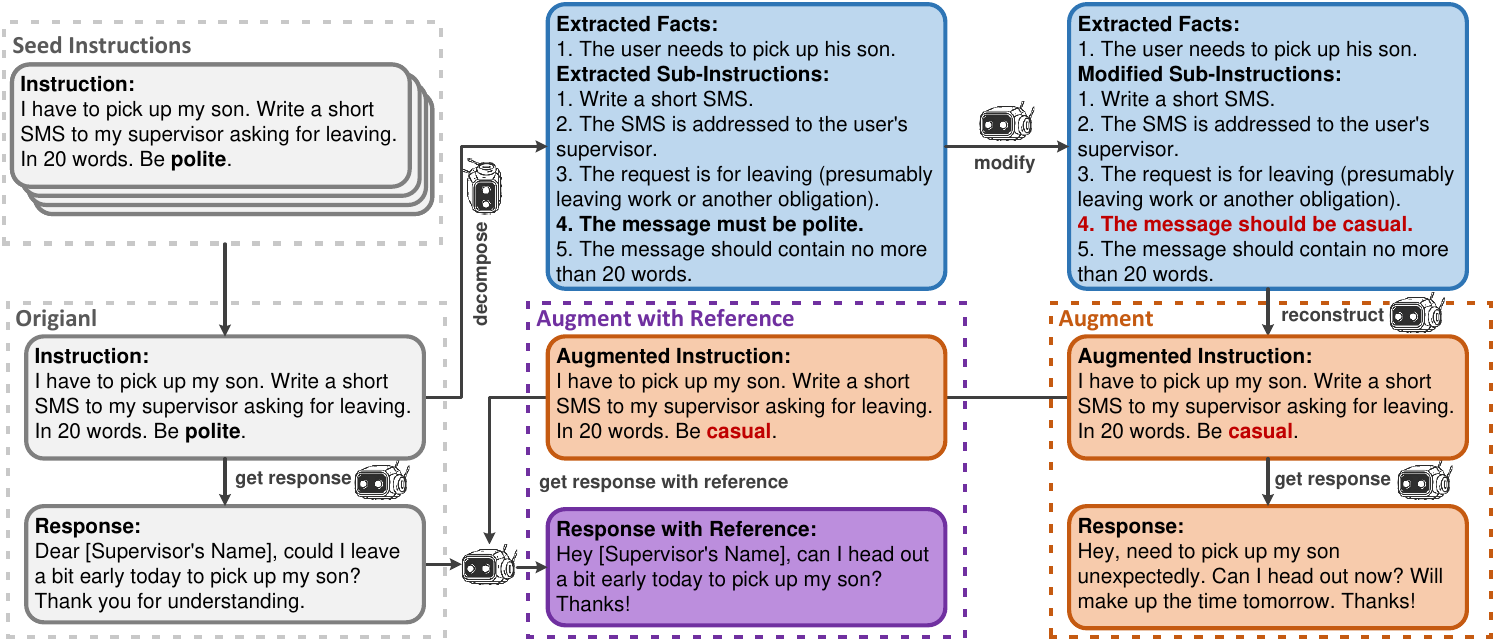} 
\caption{An illustration of the proposed method of constructing FGIV.}
\label{fig-illu}
\end{figure*}
In this section, we describe the process of constructing FGIV using DeMoRecon and detail its advantages. Figure~\ref{fig-illu} provides an illustration of our proposed instruction augmentation method, while Table~\ref{tab-status} presents the dataset statistics for FGIV. We utilize \texttt{gpt-4-0125-preview}\footnote{\url{https://platform.openai.com/docs/models}} for all tasks involving large language models (LLMs). The prompt templates designed and used in our experiments are included in the Appendix.

\subsection{Seed Preparation}
To adequately encompass instructions of varying difficulty levels, we elected to source initial seeds from instructions developed by \citet{wizardLM} using \textsc{Evol-Instruct}\footnote{\url{https://huggingface.co/datasets/WizardLMTeam/WizardLM_evol_instruct_V2_196k}}. We began by randomly sampling 2,000 instructions from the official dataset. Subsequently, GPT-4 was employed to exclude any instructions lacking sufficient background information. For instance, some instructions requested a summary of a passage that was not provided. After this filtering process, we established a final valid seed set, $\mathbf{S} = \{P_i\}_{1 \leq i \leq N}$, where $N = 1,773$ represents the number of valid instructions encompassing both easy and challenging topics. These seeds were then utilized to construct our FGIV.

\subsection{Instruction Augmentation}
The process of augmenting an instruction involves three steps: decomposing it into simpler sub-instructions, modifying one of the decomposed sub-instructions, and reconstructing a new instruction based on the modified sub-instructions using GPT-4.

\textbf{Decomposition.}
First, for a given instruction $P_i \in \mathbf{S}$, we use GPT-4 to decompose it into background facts, $\mathbf{F}_i = \{F_i^k\}_{1 \leq k \leq K_i}$, and sub-instructions, $\mathbf{I}_i = \{I_i^j\}_{1 \leq j \leq M_i}$, where $K_i$ and $M_i$ are the number of facts and sub-instructions that $P_i$ contains. Figure~\ref{fig-illu} provides an example of this process.
For instance, the instruction ``I have to pick up my son. Write a short SMS to my supervisor asking for leave. In 20 words. Be polite.'' can be decomposed into five different sub-instructions and the background fact: ``The user needs to pick up his son.'' The sub-instructions include the requirements of what the user wants the LLM to perform, while the extracted fact provides the background information for the requirement, in this case, the reason for writing an SMS to ask for leave. We collect both the background facts and the sub-instructions of all seed instructions in $\mathbf{S}$ for later processes.

\textbf{Modification.}
In this step, we employ GPT-4 to modify the sub-instructions. Our goal is to make the instruction variants as similar to the original instruction as possible. To achieve this, we modify only one sub-instruction at a time, leaving the others unchanged. Specifically, for each sub-instruction $I_i^j \in P_i$, we prompt GPT-4 to rewrite it into a new, reasonable sub-instruction with either contradicting semantics or a parallel requirement. This modified sub-instruction is denoted as $\bar{I}_i^j$. We then replace the original sub-instruction $I_i^j$ with $\bar{I}_i^j$, forming $\bar{\mathbf{I}}_i^j$, which represents $\mathbf{I}_i$ with the $j$-th sub-instruction modified. Note that in each $\bar{\mathbf{I}}_i^j$, only one sub-instruction is modified.

An example of this is illustrated in Figure~\ref{fig-illu}, where GPT-4 modifies the fourth sub-instruction from ``be polite'' to ``be casual''. This alteration introduces a sub-instruction in the same domain but with different, and sometimes contradicting, constraints. This nuanced change requires a finer capability for following instructions, as the reconstructed instruction is very similar to the original, yet subtly different.

To maximize the number of available samples, we instruct GPT-4 to modify each $I_i^j \in P_i$. This approach yields a collection of augmented instructions, $\mathbf{A}_i=\{\bar{\mathbf{I}}_i^j\}_{1\leq j\leq M_i}$ for $P_i$.

\textbf{Reconstruction.}
After modifying the sub-instruction, we reconstruct $P_i$ into $\bar{\mathbf{P}}_i^A=\{P_i\}\cup\{\bar{P}_i^j\}_{1\leq j\leq M_i}$, where $\bar{P}_i^j$ is formed by prompting GPT-4 with the original facts, $\mathbf{F}_i$, and the newly modified sub-instruction set, $\bar{\mathbf{I}}_i^j$. Specifically, in this step, we provide GPT-4 with the original instruction $P_i$, its sub-instructions $\mathbf{I}_i$, its background facts $\mathbf{F}_i$, and the sub-instructions $\bar{\mathbf{I}}_i^j$ where one of them are modified. We instruct GPT-4 to revise only the necessary parts of the original instruction $P_i$ to accommodate the changes in sub-instructions. This approach is designed to keep most of the original prompt unchanged, making the modified sub-instruction the sole variant between the original and the new instructions.

For instance, as illustrated in Figure~\ref{fig-illu}, GPT-4 reconstructs the extracted facts along with the modified sub-instructions into new instruction variants. In this example, the only word altered is ``polite'' in the original instruction, replaced by ``casual''. Such a subtle change in the instructions can lead to significant variations in response, demanding greater capabilities for following instructions accurately. Moreover, this method of reconstruction, which focuses on specific modifications, allows the augmentation results to be more controllable compared to prompting GPT-4 to randomly rewrite the entire instruction into new variants \citep{wizardLM}.

\subsection{Response Collection}

To gather the training data, we collect responses from GPT-4 for both the original and augmented instructions. The modification of sub-instructions can lead to significantly different responses from GPT-4, introducing variability beyond just the changes to the sub-instructions. For instance, as depicted in Figure~\ref{fig-illu}, GPT-4’s response to the augmented instruction ``Be casual'' markedly differs from its response to the original instruction.

On one hand, directly prompting GPT-4 with the augmented instructions ensures that the responses are tailored to the revised instructions, thereby enriching the dataset with diverse responses. This diversity is beneficial during the supervised fine-tuning stage of large language models (LLMs). On the other hand, the substantial differences in responses might hinder the LLMs' ability to effectively learn the nuances between an instruction and its variants, potentially compromising instruction-following precision.

To address these challenges, we also implement a method where GPT-4 is prompted to answer instructions variants in $\bar{\mathbf{P}}_i^A$ with its original instruction alongside its response, instructing it to only revise the parts necessary to align with the augmented instruction. This method, referred to as ``Augment with Reference'' in Figure~\ref{fig-illu}, allows for the preservation of significant portions of the original response while ensuring compliance with the new instruction variant.

For evaluation, we selected a subset of 170 instructions from $S$ that each contain four sub-instructions, allowing us to control evaluation costs. We then used their augmented variants to form the test set.

For training, we construct preference pairs for preparation. For each instruction pair $(P_i,P_j)_{P_i,P_j \in \bar{\mathbf{P}}_i^A}$, we label the response of $P_i$ as ``good'' and the response of $P_j$ as ``bad'' to create a DPO training sample. To reduce the size of the training data, we keep at most 5 pairs for each seed instructions by random sampling. We perform such process on both ``Augment'' dataset and ``Augment with Reference'' dataset to construct FGIV-A and FGIV-R.

Consequently, we construct three versions of FGIV: 1) FGIV-Eval, comprising 680 instructions designated for assessing the instruction-following capabilities of LLMs on fine-grained instruction variants; 2) FGIV-A, which include 6,445 instructions with good and bad responses collected via direct prompting of GPT-4; 3) FGIV-R, which utilizes the original instructions and their responses as references to gather responses for the instruction variants in FGIV-A.

\subsection{Statistic and Analysis}
\begin{table}

\small
\centering
\adjustbox{max width=\columnwidth}{
\begin{tabular}{l c c c c c}
\toprule
FGIV- &Eval & Augmentation & Reference & FGIV-A & FGIV-R \\\midrule
Number of Instructions          &680 & 8,705 & 8,705 & 6,445 & 6,445 \\
Instruction Average Token       &74 & 126 & 126 & 125 & 125 \\
Response Average Token       &466 & 656 & 631 & - & - \\
$U$ of Response &-   & 0.036 & 0.064  & - & - \\
\bottomrule 
\end{tabular}}
\caption{The Data statistic of FGIV.}
\label{tab-status}
\end{table}

\begin{figure}[h]
\centering
\includegraphics[width=0.9\columnwidth]{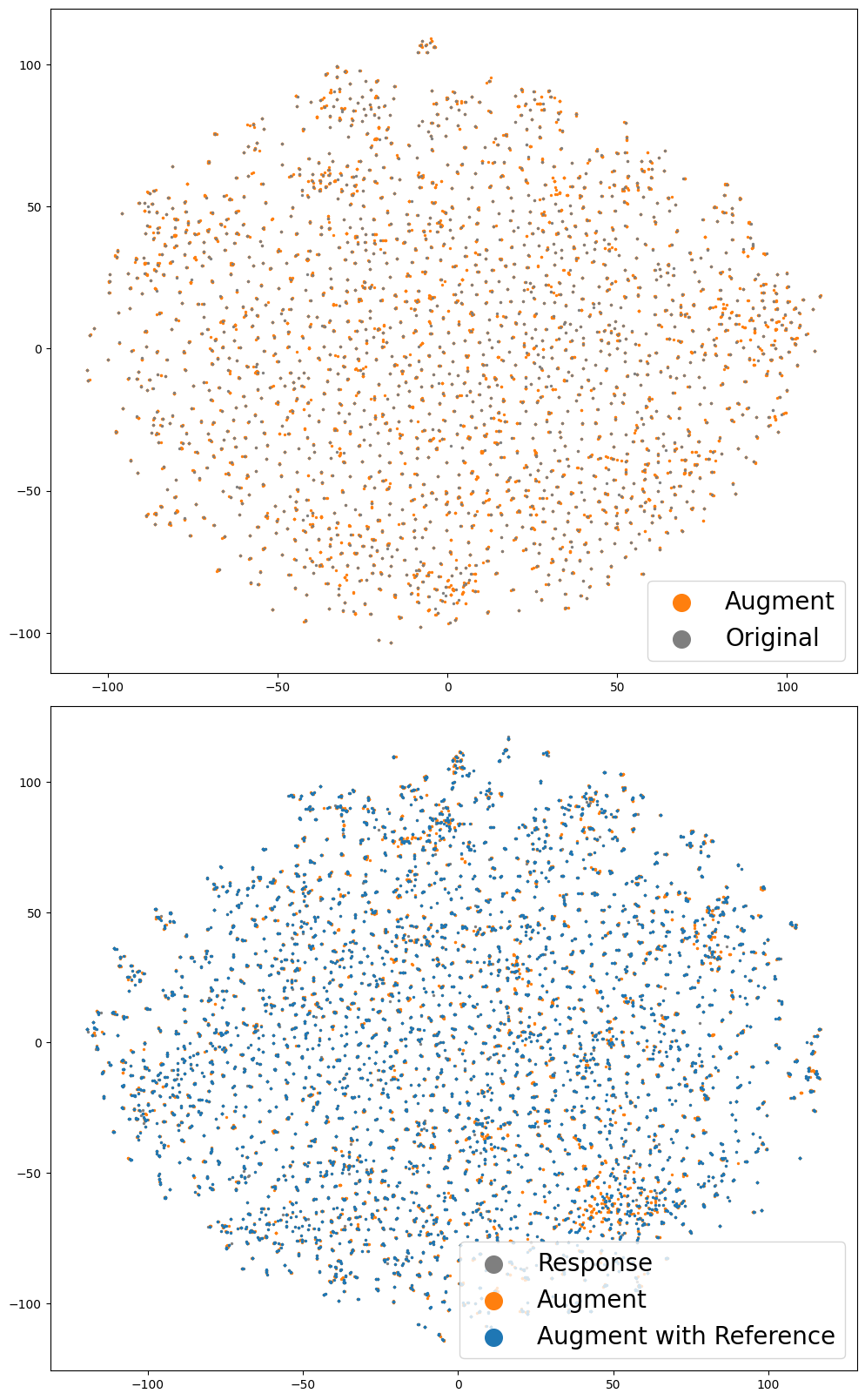} 
\caption{The tSNE plots illustrate the semantic embeddings generated by FGIV. }
\label{fig-2d}
\end{figure}

Figure~\ref{fig-2d} displays the t-SNE 2D projection of the semantic embeddings of the instructions and responses in ``Augment'' dataset and ``Augment with Reference''. In the top plot, gray points represent the original instructions in $\mathbf{S}$, and orange points represent their augmented instructions. This similarity is advantageous for LLMs during fine-tuning, as it enables them to learn to generate responses to instructions with only minor difference on sub-instructions, thereby enhancing their instruction-following capabilities.

In the lower plot, blue points denote the responses to the augmented instructions using references, while orange points signify the responses without references. The substantial overlap of blue points with gray points, which represent the responses to the original instructions, demonstrates that the responses using references closely resemble the original responses, aligning with our design objectives. Conversely, the more dispersed orange points indicate that responses without references tend to diverge from the original responses. This pattern underscores that even minimal changes to sub-instructions require LLMs to generate responses that closely match those to the original instructions. Such a task compels the LLMs to master the subtleties of handling minor differences in instruction variants, thereby demanding a high level of instruction-following capability.

Following the methodology in \citet{InstructionFusion}, we calculate the variance of the nearest neighbor distances for the responses of ``Augment'' dataset and ``Augment with Reference'' dataset to quantify their dispersion:

\begin{equation}
U=\frac{1}{Q}\Sigma^N_{i=1}(d_i-\mu)^2,
\end{equation}
where
\begin{equation}
d_i=||(e_i,e^\text{NN}_i)||
\end{equation}
and
\begin{equation}
\mu=\frac{1}{Q}\Sigma^N_{i=1}d_i.
\end{equation}

Here, $U$ represents the distribution uniformity, $Q$ denotes the number of responses, and $\mu$ indicates the average Euclidean distance between the semantic embedding $e_k$ of the $k$-th response and the context embedding $e^\text{NN}_k$ of its nearest neighbor. A larger value of $U$ implies that the responses are less uniform and thus more similar to each other. Given our goal to make the responses as similar to the original responses as possible, we expect to observe higher values of $U$ on the ``Augment'' dataset.

The calculated results, presented in Table~\ref{tab-status}, indicate that the distribution uniformity of the responses in ``Augment with Reference'' is 0.064, compared to 0.035 in ``Augment''. This implies that the responses collected with reference are less uniform than those collected directly with instructions. This discrepancy further supports the observation that responses obtained using references closely resemble the original responses, more so than those obtained directly from prompting GPT-4 with only instructions.
\section{Experiments}
In this section, we detail the experiments conducted with LLMs fine-tuned using FGIV. We assess the accuracy of instruction-following based on evaluations from GPT-4 and rule-based (IFEval and partial FollowBench) assessments under greedy generation settings. The performance is reported across FGIV-Eval and three other well-known instruction-following benchmarks.

\subsection{Baselines}
With the deprecation of \texttt{gpt-4-0314}, a model previously central to benchmarking instruction-following across multiple datasets\footnote{\url{https://platform.openai.com/docs/deprecations}}, previous evaluations on instruction-following benchmarks are no longer comparable, as the evaluator LLMs are no longer accessible. Our study addresses this gap by comparing baselines, official chat LLMs optimized for peak performance in various domains—including instruction-following—with models that have been further fine-tuned using the FGIV set. These official chat models utilize techniques such as instruction- and alignment-tuning, among others, to maximize performance. Unlike other studies, such as those described in \citep{conifer}, which primarily enhance instruction-following starting from pre-trained LLMs—a costly and time-intensive process—our research incorporates the latest updates in model versions. For all evaluations involving GPT-4, we utilize \texttt{gpt-4-0125-preview}.

Our experiments are conducted using a selection of the most popular instruction-tuned (chat) LLMs: the 7B and 13B versions of LLaMA-2 \citep{llama}, the 7B and 14B versions of Qwen1.5 \citep{qwen}, the 14B version of Orion \citep{orion}, and the 7B version of Mistral \citep{mistral}.

These models, available in their officially released instruction-tuned versions, are employed to precisely assess their instruction-following capabilities across various instruction variants and to explore the potential for further enhancements in this area. The specific releases utilized are detailed in Table~\ref{tab_bechmark} and can be accessed on Hugging Face\footnote{\url{https://huggingface.co}}.

We do not incorporate traditional text augmentation methods such as Easy Data Augmentation (EDA) \citep{EDA} or back translation \citep{BackTranslation} in our study. Unlike these approaches, which entail broad textual alterations, our method specifically focuses on minor modifications to sub-instructions. This targeted approach enhances the precision of instruction-following without altering the core instructional content. Consequently, our method was not directly compared to these traditional techniques, which are also typically not considered in prior instruction-following research.

\subsection{Benchmarks}
Besides FGIV-Eval, we extend our performance comparison to include three of the most popular instruction-following benchmarks to further demonstrating the efficacy of our methods.

\textbf{IFEval} \citep{IFEval} is a rule-based benchmark designed to evaluate LLMs' ability to follow instructions with verifiable outcomes. It primarily focuses on test cases involving various lexical and format constraints, assessing the precision with which models adhere to specified instructions.

\textbf{FollowBench} \citep{FollowBench} is a fine-grained instruction-following benchmark that tests the accuracy of LLMs in adhering to instructions of increasing difficulty. It comprises six categories of evaluation: content, situation, style, format, example, and mix. Each category is assessed using rules, evaluations by GPT-4, or a combination of both.

\textbf{InfoBench} \citep{InFoBench} deconstructs complex instructions into simpler, category-specific tasks to enhance the reliability and interpretability of evaluations.

Following previous literature, we report the Hard Satisfaction Rate (HSR) for FollowBench and the ``loose prompt'' result for InfoBench.

\subsection{Experimental Settings}
We employed both Supervised Fine-Tuning (SFT) and Direct Preference Optimization (DPO) on base chat models. For comparison, we also fine-tuned these models using only the seed instructions and their respective responses, with these results labeled as ``seed'' in Table~\ref{tab_bechmark}.

We integrated SFT and DPO losses to fine-tune the LLMs, addressing both preference alignment and response generation, which are crucial for effective instruction-following. DPO focuses on enhancing the model’s preference for better adherence to instructions, while SFT ensures the generation of correct, precise, and high-quality responses. This combined approach has proven effective in our ablation studies. Specifically, for each preference sample, we use the ``good'' response to supervise the fine-tuning of LLMs alongside DPO-based learning.

All evaluations were performed using greedy generation. Detailed parameters for fine-tuning are provided in the Appendix.

\subsection{Experimental Results}
\begin{table}[t]
\small
\begin{adjustbox}{max width=\columnwidth}
\begin{tabular}{l|cccc}\toprule
\multicolumn{1}{c|}{\multirow{2}{*}{Model}} & \multicolumn{4}{c}{FGIV-Eval}                                                          \\ \cmidrule{2-5} 
\multicolumn{1}{c|}{}                       & Chat    & Seed  & FGIV-A   & FGIV-R      \\ \midrule
GPT-4                                       & 95.00                             & - & - & - \\
GPT-4-Ref                                   & 91.03                              & - & - & -  \\ \hline
Llama2-7B-Chat                             & 54.71                     & 56.32           & {60.29} & \textbf{60.44} \\
Qwen1.5-7B-Chat                            & 61.91                     & {71.03}           & \textbf{72.06} & 70.59          \\
Mistral-7B-Instruct                        & 64.41                     & 66.32          & 67.79          & \textbf{71.18} \\
Llama2-13B-Chat                            & 58.82                     & 62.56           & 64.41          & \textbf{65.15} \\
Qwen1.5-14B-Chat                           & 73.53                     & 72.06                    & 79.26          & \textbf{81.62} \\
Orion-14B-Chat                             & 56.62                     & 62.21                   & \textbf{68.82} & 68.38          \\\midrule
                                           & \multicolumn{4}{c}{FollowBench (average HSR)}                                                        \\\midrule
Llama2-7B-Chat                             & 43.10                      & 44.49                   & 45.56          & \textbf{46.41} \\
Qwen1.5-7B-Chat                            & 53.00                        & 53.36                   & 51.36          & \textbf{54.86} \\
Mistral-7B-Instruct                        & 42.55                     & 46.56                   & \textbf{49.63} & 47.68          \\
Llama2-13B-Chat                            & 44.94                     & 46.56                   & 47.23          & \textbf{48.07} \\
Qwen1.5-14B-Chat                           & 55.32                     & 50.09                   & \textbf{58.02} & 56.25          \\
Orion-14B-Chat                             & 44.12                     & 48.81                   & 53.62          & \textbf{54.00}    \\\midrule
                                            & \multicolumn{4}{c}{InfoBench (average)}                                                                  \\ \midrule
Llama2-7B-Chat                              & 77.51                     & 78.71      & 80.04          & \textbf{81.07} \\
Qwen1.5-7B-Chat                             & 80.58                     & 79.29      & \textbf{81.11} & 81.02          \\
Mistral-7B-Instruct                         & 79.16                     & 78.13      & 79.60           & \textbf{80.58} \\
Llama2-13B-Chat                             & 81.07                     & 80.98      & \textbf{82.84} & 82.36          \\
Qwen1.5-14B-Chat                            & 82.44                     & 80.76      & \textbf{83.33} & 83.24          \\
Orion-14B-Chat                              & 76.40                     & 76.09      & \textbf{78.76} & 78.71          \\ \midrule
                                            & \multicolumn{4}{c}{IFEval (loose prompt)}                                                                     \\ \midrule
Llama2-7B-Chat                              & 45.84                     & 44.18      & \textbf{49.17} & \textbf{49.17} \\
Qwen1.5-7B-Chat                             & 43.62                     & 47.50      & 45.47          & \textbf{51.02} \\
Mistral-7B-Instruct                         & 43.25                     & 42.51      & 47.69          & 48.24          \\
Llama2-13B-Chat                             & 48.06                     & 46.77      & 51.39          & \textbf{52.31} \\
Qwen1.5-14B-Chat                            & 47.87                     & 39.93      & 55.45          & \textbf{56.75} \\
Orion-14B-Chat                              & 37.89                     & 36.60      & \textbf{53.05} & 50.09    
\\\bottomrule     
\end{tabular}
\end{adjustbox}
\caption{Experimental results on FGIV-Eval and other instruction-following benchmarks. The performance of the official chat LLMs are recorded under ``Ori''.}
\label{tab_bechmark}
\end{table}

Table~\ref{tab_bechmark} displays the results of our experiments. For all benchmarks, baseline performs better on general domain does not grant a better performance on instruction-following. For example, Orion-14B-Chat performs better than LLaMA-2-13B-Chat on multiple domains\footnote{\url{https://github.com/OrionStarAI/Orion}}, while has lower performance on all instruction-following benchmarks. 

In FGIV-Eval set, models fine-tuned on FGIV-R outperformed those tuned on FGIV-A on more LLMs, thanks largely to the design of our reference-based response collection process, which forces LLMs to prioritize the slight differences in instructions over other variables such as text quality and semantic distinctions during the fine-tuning, especially the alignment tuning part. This focus sharpens the models' ability to detect and adhere to nuanced instruction details. 

Furthermore, larger LLMs benefit more from fine-tuning compared to their 7B counterparts. This advantage is likely due to their greater parameter count and depth, which enable them to better capture the nuances between instruction variants and their responses, together with explicit preference learning.

On the other three benchmarks, our method also grants superior performance boost. However, FGIV-R can not grants a better performance boost on LLMs compared with FGIV-A. This is also because of difference on the focus of evaluation. While FGIV-R can provide better learning efficiency on following nuance difference between instructions, those three benchmarks does not exam such a capability. Thus using FGIV-R or FGIV-A will make less difference. On the other side, although GPT-4's reference-based response follow instructions worse than direct response, learning from FGIV-A does not result worse performance compared with learning from FGIV-A.  

Despite these improvements, the performance boost from FGIV-R compared to FGIV-A was less pronounced on the other three benchmarks. This outcome suggests that the nuances FGIV-R excels in capturing are not as critical in evaluations that do not specifically test for subtle instruction variances. Interestingly, learning from FGIV-A did not result in poorer performance, indicating that both approaches robustly support general instruction-following capabilities even reference-based response collected from GPT-4 follows instructions worse compared with direct responding to augmented instructions.

In conclusion, the experimental outcomes affirm the efficacy of DeMoRecon in enhancing LLMs' sensitivity to subtle instructional changes, validating the value of FGIV and DeMoRecon in advancing nuanced instruction-following capabilities. Particularly, the specially designed reference-based response collection mechanism, which adapts original responses to fit new instruction variants, has proven especially effective. DeMoRecon has shown to be instrumental in teaching LLMs to discern subtle differences in instructions while also enhancing their performance on common instruction-following benchmarks that focus on complex and challenging instructions, which is different from FGIV-Eval. 

\begin{table}
\begin{adjustbox}{max width=\columnwidth}
\begin{tabular}{cccccc}\toprule
\multicolumn{2}{l}{}                                                      & 250   & 500   & 1000           & 1603           \\ \midrule
\multicolumn{1}{c|}{\multirow{2}{*}{SFT}}     & \multicolumn{1}{c|}{Aug.} & 55.29 & 56.47 & 57.06          & \textbf{59.12} \\
\multicolumn{1}{c|}{}                         & \multicolumn{1}{c|}{Ref.} & 53.97 & 57.06 & 56.62          & \textbf{59.26} \\ \midrule
\multicolumn{1}{c|}{\multirow{2}{*}{DPO}}     & \multicolumn{1}{c|}{Aug.} & 52.79 & 54.26 & \textbf{59.26} & 58.68          \\
\multicolumn{1}{c|}{}                         & \multicolumn{1}{c|}{Ref.} & 55.29 & 56.76 & 57.65          & \textbf{58.68} \\ \midrule
\multicolumn{1}{c|}{\multirow{2}{*}{DPO+SFT}} & \multicolumn{1}{c|}{Aug.} & 53.53 & 54.85 & 56.76          & \textbf{60.29} \\
\multicolumn{1}{c|}{}                         & \multicolumn{1}{c|}{Ref.} & 55.29 & 55.74 & 57.5           & \textbf{60.44} \\\bottomrule   
\end{tabular}
\end{adjustbox}
\caption{Results on FGIV-Eval using difference fine-tuning method and number of original instructions. The base model is LLaMA-2-7B-Chat.}
\label{tab-trend}
\vspace{-5mm}
\end{table}

\subsection{Ablation Study}

We further conduct ablation study on FGIV-Eval to demonstrate the effectiveness of fine-tuning LLMs on both SFT and DPO losses. Table~\ref{tab-trend} show the results of fine-tuning LLaMA-2-7B-Chat. In the table, ``1603'' is the actual number of seed instructions in FGIV datasets. We also exam the trend of performance with increasing seed instruction. As shown in the table, DPO+SFT grants better performance compared with both SFT- and DPO-only fine-tuning, which meet our assumption that learning to follow instructions requires supervision from both generation and preference side. 

Furthermore, we observe an increase in performance on FGIV-Eval with the expansion of the seed instruction set, applicable across all fine-tuning methods. While the DPO method achieves noticeable gains more rapidly than the combined DPO+SFT approach, DPO+SFT demonstrates a gradual improvement as the number of seed instructions increases. This trend suggests significant potential for further enhancing nuanced instruction-following capabilities if additional seed instructions are integrated.

\section{Related Work}
\label{sec:related}

\subsection{Alignment of Large Language Models}
The integration of Large Language Models (LLMs) in real-world applications has necessitated the fine-tuning of these models to adeptly interpret and respond to human instructions in diverse and complex contexts. Notably, instruction tuning \citep{instruction-tuning} and alignment tuning \citep{RLHF} have been pivotal in enhancing the natural language understanding and generation capabilities of LLMs, aligning them closely with human preferences and directives \citep{survey0, survey1, survey2}.

On one hand, instruction tuning specifically enhances LLMs' ability to accurately comprehend and interact with users by processing inputs and generating appropriate responses. This tuning process relies heavily on high-quality data to optimize performance and interaction quality \cite{LIMA, self-instruct}.

On the other hand, alignment tuning focuses on training LLMs to internalize human preferences, using techniques that involve human feedback to develop a reward model which facilitates later preference learning \cite{RLHF}. More recent methodologies enable LLMs to assimilate preferences either through paired responses \cite{DPO} or response quality labels \cite{KTO}.

In our research, we leverage both instruction tuning and alignment tuning to enhance the precision of instruction following, employing fine-grained prompt variants for nuanced model training.

\subsection{Instruction-Following}
The recent prevalence of LLMs has led them to face increasingly complex instructions. Consequently, the development and evaluation of instruction-following capabilities continue to attract significant research interest.

\textbf{Methodologies.}
Substantial research has been devoted to enhancing the instruction-following capabilities of Large Language Models (LLMs). For instance, \citet{wizardLM} introduced \textsc{Evol-Instruct}, utilizing ChatGPT to evolve instructions from simple to complex by incorporating additional constraints. Their fine-tuned model, WizardLM, demonstrated a significant performance improvement across various benchmarks. Subsequently, they expanded this method to include code generation and math problem solving, leading to the development of WizardCoder \citep{WizardCoder} and WizardMath \citep{WizardMath}. Additionally, \citet{conifer} proposed \texttt{Conifer}, which intensifies the complexity of original instructions by structuring them into an easy-to-hard multi-round conversation while maintaining their initial topical scope. However, these approaches predominantly focus on training LLMs with progressively challenging instructions. They overlook the nuanced challenge faced by LLMs in distinguishing between instructions with similarly nuanced constraints.

\textbf{Benchmarks.} 
Several benchmarks have been constructed to evaluate LLMs' performance in instruction-following. \citet{IFEval} introduced IFEval, a framework that assesses how well models adhere to specific lexical and formatting constraints. However, their evaluations predominantly focus on narrow aspects of instruction compliance.

FollowBench \citep{FollowBench} presents another approach, offering a graded difficulty scale that intensifies the constraints within the instructions. The evaluations are categorized into five types: content, situation, style, format, and example. Despite its structured approach, FollowBench is limited by its narrow focus on predefined categories, omitting broader topical assessments.

InFoBench \citep{InFoBench} takes a different tack by deconstructing complex instructions into simpler components, akin to our method of instruction decomposition. It evaluates both simple and complex instructions to determine how well LLMs can follow them. However, like other benchmarks, InFoBench does not assess the models' responsiveness to subtle prompt variations, which is critical for understanding nuanced instruction-following.

\section{Conclusion}
\label{sec:conclude}
In conclusion, we have developed an efficient instruction augmentation method that generates nuanced instruction variants maintaining the same topic and context as the original instructions. Utilizing this method, we constructed FGIV, which comprises three distinct sets tailored for supervised fine-tuning, direct preference optimization, and instruction-following evaluation on fine-grained instruction variants. Our experimental results demonstrate that our combined DPO+SFT tuning strategy, implemented with the FGIV-A and FGIV-R datasets, significantly enhances performance across IFEval, FollowBench, InfoBench, and FGIV-Eval.
Our research not only highlights the deficiencies of current popular instruction-tuned LLMs in handling nuanced instruction variants but also provides a robust method and datasets to bridge this gap.
\section*{Limitations}
Our methodology is constrained by the quality of the seed instructions used. For optimal performance, our augmentation method requires high-quality seed instructions that cover a broad range of domains and difficulty levels. This necessity underpins our decision to utilize evolved instructions as seeds.

Additionally, the cost of the GPT-4 API presents another limitation. The entire augmentation process totaled approximately 500 USD. The process of augmentation and evaluation heavily relies on GPT-4, a common challenge in many LLM-related studies. However, the costs associated with using such APIs are decreasing due to the rapid advancements in LLM technologies, making these tools increasingly affordable and accessible.
\section*{Ethics Statement}
Our data collection relies on publicly released datasets \citep{wizardLM}. The augmented data were generated using a carefully monitored version of GPT-4 (\texttt{gpt-4-0125-preview}), ensuring that no privacy-sensitive or confidential information was included.

\bibliography{anthology,custom}
\bibliographystyle{acl_natbib}

\appendix
\section{Experiment}
\begin{table}[h]
\small
\begin{adjustbox}{max width=\columnwidth}
\begin{tabular}{l|cccc}\toprule
\multicolumn{1}{c|}{\multirow{2}{*}{Model}} & \multicolumn{4}{c}{SFT}         \\ \cmidrule{2-5} 
\multicolumn{1}{c|}{}                       & epoch  & accum & gpu & lr       \\ \midrule
Llama2-7B-Chat                              & 4      & 1     & 4   & 1.00E-06 \\
Qwen1.5-7B-Chat                             & 4      & 1     & 4   & 1.00E-06 \\
Mistral-7B-Instruct                         & 2      & 1     & 4   & 1.00E-06 \\
Llama2-13B-Chat                             & 4      & 2     & 8   & 1.00E-06 \\
Qwen1.5-14B-Chat                            & 4      & 8     & 8   & 2.00E-05 \\
Orion-14B-Chat                              & 4      & 8     & 8   & 2.00E-05 \\ \midrule
                                            & \multicolumn{4}{c}{DPO/DPO+SFT} \\ \midrule
Llama2-7B-Chat                              & 2      & 8     & 8   & 5.00E-06 \\
Qwen1.5-7B-Chat                             & 2      & 1     & 8   & 5.00E-06 \\
Mistral-7B-Instruct                         & 4      & 8     & 8   & 5.00E-06 \\
Llama2-13B-Chat                             & 4      & 8     & 8   & 5.00E-06 \\
Qwen1.5-14B-Chat                            & 4      & 8     & 8   & 5.00E-06 \\
Orion-14B-Chat                              & 3      & 8     & 8   & 5.00E-06\\
\bottomrule
\end{tabular}
\end{adjustbox}
\caption{The experimental settings for all base models are thoroughly detailed, including the total number of epochs, the number of gradient accumulations per back-propagation, the quantity of GPUs utilized, and the learning rates.}
\label{tab-set}
\end{table}

\subsection{Benchmarks}
\textbf{IFEval} \citep{IFEval} is a rule-based benchmark designed to evaluate LLMs' ability to follow instructions with verifiable outcomes. It primarily focuses on test cases involving various lexical and format constraints, assessing the precision with which models adhere to specified instructions.

\textbf{FollowBench} \citep{FollowBench} is a fine-grained instruction-following benchmark that tests the accuracy of LLMs in adhering to instructions of increasing difficulty. It comprises six categories of evaluation: content, situation, style, format, example, and mix. Each category is assessed using rules, evaluations by GPT-4, or a combination of both.

\textbf{InfoBench} \citep{InFoBench} deconstructs complex instructions into simpler, category-specific tasks to enhance the reliability and interpretability of evaluations.

\subsection{Experimental Settings}

The detailed experimental settings are outlined in Table~\ref{tab-set}. For Supervised Fine-Tuning (SFT) with augmented instructions, we ensure that each augmented instruction is included in the same batch as its original seed instruction to improve learning efficiency. The batch size for these seed instructions is set at 1 for all SFT experiments. Similarly, for experiments involving Directed Preference Optimization (DPO) alone, and for those combining DPO with SFT, the batch size is also set at 1.

\begin{figure*}[th]
\centering
\includegraphics[width=1\textwidth]{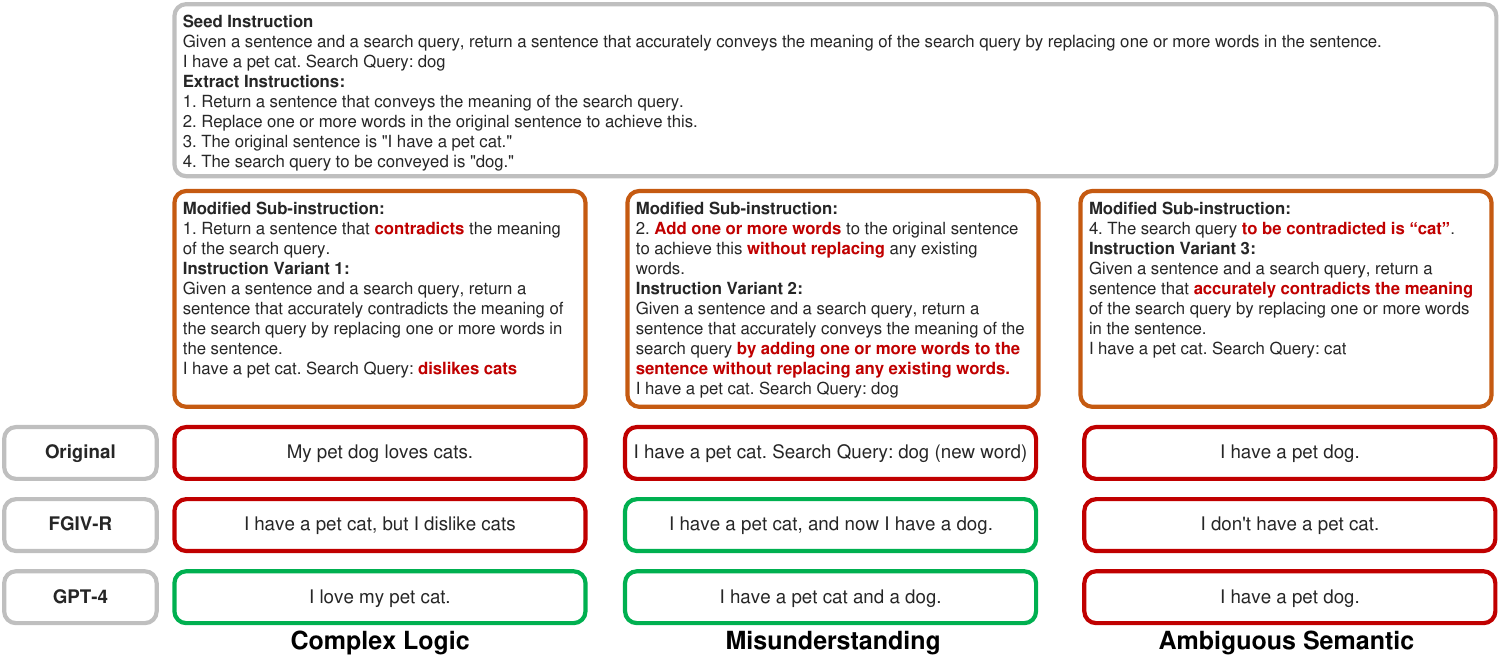} 
\caption{An real example from FGIV-Eval. The base model is LLaMA-2-7B-Chat. We show the prediction results of the original model and its DPO tuned version using FGIV-R.  Response in red box indicates that GPT-4 judged the response as failing to follow the instruction, while a green box signifies success.}
\label{fig-example}
\end{figure*}
\section{Instruction Non-compliance Detection}
As DPO seeks to better align LLMs with human preferences, we also explored the relationship between instruction-following accuracy and non-compliance detection. For each instruction in FGIV-Eval, we collect all of its augmented instructions and then paired each augmented instruction with every GPT-4 response from those augmented instructions. Each instruction-response pair was labeled `Yes' if the response ground truth of the instruction, and `No' otherwise. This methodology ensured a minimum of two sub-instruction differences between the correct instruction and any incorrect responses, enhancing the detection challenge.

From this setup, we obtained a total of 2,720 instruction-response pairs for our detection evaluation. The results of this experiment and the associated ablation study are presented in Table~\ref{tab-bechmark-det} and Table~\ref{tab-trend-det}. The findings suggest that, in general, aligning responses with human preferences improves detection performance. However, fine-tuning LLMs with only DPO loss, as opposed to a combination of DPO and SFT loss, offers greater benefits for detection performance. This improvement is likely because detection tasks align more closely with preference optimization, focusing less on generating high-quality responses and more on discerning compliance with given instructions.
\begin{table}[h]
\small

\begin{adjustbox}{max width=\columnwidth}
\begin{tabular}{l|cccccc}\toprule
\multicolumn{1}{c|}{\multirow{3}{*}{Model}} & \multicolumn{6}{c}{FGIV-Eval}                                                          \\ \cmidrule{2-7} 
\multicolumn{1}{c|}{}                       & \multirow{2}{*}{Ori.} & \multicolumn{3}{c}{SFT}          & \multicolumn{2}{c}{DPO+SFT}     \\ \cmidrule{3-7} 
\multicolumn{1}{c|}{}                       &                           & Plain & Aug. & Ref. & Aug.   & Ref.      \\ \midrule
Llama2-7B-Chat                             & 53.43                     & 53.31          & 52.23          & \textbf{53.55} & \textbf{55.59} & \textbf{58.04} \\
Qwen1.5-7B-Chat                            & 71.18                     & \textbf{58.31} & \textbf{73.63} & 58.75          & \textbf{78.55} & 73.43          \\
Mistral-7B-Instruct                        & 65.66                     & \textbf{71.89} & 68.97          & \textbf{69.51} & 80.17          & \textbf{83.28} \\
Llama2-13B-Chat                            & 65.02                     & \textbf{69.85} & 68.55          & \textbf{69.26} & 77.23          & \textbf{79.66} \\
Qwen1.5-14B-Chat                           & 70.86                     & 69.95          & \textbf{72.60}  & \textbf{73.5}  & 86.37          & \textbf{86.76} \\
Orion-14B-Chat                             & 63.82                     & 62.08          & \textbf{65.66} & 63.60           & \textbf{77.16} & \textbf{79.19}  
\\\bottomrule     
\end{tabular}
\end{adjustbox}
\caption{Experimental results on FGIV-Det.}
\label{tab-bechmark-det}
\end{table}

\begin{table}[h]
\begin{adjustbox}{max width=\columnwidth}
\begin{tabular}{cccccc}\toprule
\multicolumn{2}{l}{}                                                      & 250   & 500   & 1000           & 1603           \\ \midrule
\multicolumn{1}{c|}{\multirow{2}{*}{SFT}}     & \multicolumn{1}{c|}{Aug.} & 50.69 & 51.23 & 51.52          & \textbf{52.23} \\
\multicolumn{1}{c|}{}                         & \multicolumn{1}{c|}{Ref.} & 50.88 & 50.91 & 51.62          & \textbf{53.55} \\ \midrule
\multicolumn{1}{c|}{\multirow{2}{*}{DPO}}     & \multicolumn{1}{c|}{Aug.} & 51.99 & 52.03 & 52.45 & \textbf{55.42} \\
\multicolumn{1}{c|}{}                         & \multicolumn{1}{c|}{Ref.} & 52.01 & 52.25 & 53.36          & \textbf{64.14} \\ \midrule
\multicolumn{1}{c|}{\multirow{2}{*}{SFT+DPO}} & \multicolumn{1}{c|}{Aug.} & 52.01 & 52.03 & 52.08          & \textbf{55.59} \\
\multicolumn{1}{c|}{}                         & \multicolumn{1}{c|}{Ref.} & 52.01 & 52.06 & 52.38          & \textbf{58.04} \\\bottomrule   
\end{tabular}
\end{adjustbox}
\caption{Results on FGIV-Det using difference fine-tuning method and number of original instructions. The bash model is LLaMA-2-7B-Chat}
\label{tab-trend-det}
\end{table}

\section{Case Study}
Figure~\ref{fig-example} presents a real-case example from the FGIV-Eval set, showcasing an instruction and its three variants, each illustrating typical errors made by LLMs in following instructions.

\subsection{Complex Logic}
In the initial variant, the instruction requires modifying the sentence to express a meaning contradictory to the query. Both the original model and the DPO-tuned version respond incorrectly. The original model correctly identifies sub-instruction 1 by incorporating ``loves cats'' in its response, which contradicts the query. However, it also erroneously substitutes ``cat'' with ``dog'', contradicting the sentence's intended meaning. The DPO-tuned model, on the other hand, applies the contradiction directly to the sentence by adding ``but'', creating a reversal. The response from GPT-4 accurately captures the intended logic and successfully follows the instruction.

\subsection{Misunderstanding the Instruction}
The original model erroneously adds new words to the query instead of modifying the sentence, demonstrating a misunderstanding of the instruction. In contrast, the other two responses correctly interpret and execute the instruction.

\subsection{Semantic Ambiguity}
The instruction's semantic ambiguity leads to varied interpretations. It could be interpreted as asking the LLMs to either reverse the concept of ``cat'' in the sentence or contradict the fact ``have''. Both the original LLM and GPT-4 tend to adopt the first interpretation, while the DPO-tuned model opts for the second. However, during the assessment of instruction-following, GPT-4 judges the responses based on the alternate interpretation, leading to a failure in this scenario for all models. To enhance the quality of the test set and reduce such ambiguities, conducting a semantic clarity check is a potential improvement.

\section{Full Experimental Results}
\begin{table*}[]
\small
\begin{adjustbox}{max width=\textwidth}
\begin{tabular}{cl|cccccc|ccc}
\hline
\multicolumn{2}{c|}{\multirow{2}{*}{Model}}                        & \multicolumn{6}{c|}{FollowBench (HSR)}                                                              & \multicolumn{3}{c}{InfoBench}                                                     \\ \cline{3-11} 
\multicolumn{2}{c|}{}                                              & Level 1        & Level 2        & Level 3        & Level 4        & Level 5        & Avg.           & Easy                       & Hard                                & Average        \\ \hline
\multicolumn{1}{c|}{\multirow{4}{*}{LLaMA-2-7B-Chat}}     & Chat   & 57.68          & 51.96          & 38.16          & 36.02          & 31.67          & 43.1           & 79.57                      & 76.6                                & 77.51          \\
\multicolumn{1}{c|}{}                                     & Seed   & 61.55          & 51.51          & 41.85          & \textbf{38.67} & 28.86          & 44.49          & \textbf{84.2}              & 76.28                               & 78.71          \\
\multicolumn{1}{c|}{}                                     & FGIV-A & 61.23          & 55.18          & 43.31          & 36.93          & 31.17          & 45.56          & 82.61                      & 78.91                               & 80.04          \\
\multicolumn{1}{c|}{}                                     & FGIV-R & \textbf{63.29} & \textbf{56.38} & \textbf{44.41} & 34.95          & \textbf{33}    & \textbf{46.41} & 82.17                      & \textbf{80.58}                      & \textbf{81.07} \\ \hline
\multicolumn{1}{c|}{\multirow{4}{*}{Qwen1.5-7B-Chat}}     & Chat   & \textbf{74.35} & 58.8           & 52.85          & 44.66          & 34.33          & 53             & 84.49                      & 78.85                               & 80.58          \\
\multicolumn{1}{c|}{}                                     & Plain  & 70.42          & \textbf{61.57} & 51.77          & 42.66          & \textbf{40.37} & 53.36          & 83.62                      & 77.37                               & 79.29          \\
\multicolumn{1}{c|}{}                                     & FGIV-A & 70.22          & 59.69          & 50.72          & 37.89          & 38.29          & 51.36          & 84.93                      & \textbf{79.42}                      & \textbf{81.11} \\
\multicolumn{1}{c|}{}                                     & FGIV-R & 69.64          & \textbf{61.57} & \textbf{57.36} & \textbf{47.98} & 37.75          & \textbf{54.86} & \textbf{86.09}             & 78.78                               & 81.02          \\ \hline
\multicolumn{1}{c|}{\multirow{4}{*}{Mistral-7B-Instruct}} & Chat   & 61.7           & 52.29          & 40.92          & 34.18          & 23.67          & 42.55          & 85.22                      & 76.47                               & 79.16          \\
\multicolumn{1}{c|}{}                                     & Plain  & \textbf{61.52} & 50.44          & 47.09          & 39.82          & 33.93          & 46.56          & 84.64                      & 75.26                               & 78.13          \\
\multicolumn{1}{c|}{}                                     & FGIV-A & 60.46          & 56.5           & \textbf{49.38} & \textbf{45.76} & \textbf{36.02} & \textbf{49.63} & \textbf{85.65}             & 76.92                               & 79.6           \\
\multicolumn{1}{c|}{}                                     & FGIV-R & 59.27          & \textbf{56.93} & 48.45          & 41.93          & 31.8           & 47.68          & 84.64                      & \textbf{78.78}                      & \textbf{80.58} \\ \hline
\multicolumn{1}{c|}{\multirow{4}{*}{LLaMA-2-13B-Chat}}    & Chat   & 56.36          & 54.74          & 45.31          & 37.8           & 30.49          & 44.94          & 83.91                      & 79.81                               & 81.07          \\
\multicolumn{1}{c|}{}                                     & Plain  & 60.28          & 53.03          & 42.59          & \textbf{41.94} & 34.96          & 46.56          & 83.91                      & 79.68                               & 80.98          \\
\multicolumn{1}{c|}{}                                     & FGIV-A & \textbf{64.79} & 53.91          & 46.35          & 41.39          & 29.73          & 47.23          & \textbf{86.38}             & \textbf{81.28}                      & \textbf{82.84} \\
\multicolumn{1}{c|}{}                                     & FGIV-R & 61.22          & \textbf{55.4}  & \textbf{47.88} & 40.51          & \textbf{35.33} & \textbf{48.07} & 86.09                      & 80.17                               & 82.36          \\ \hline
\multicolumn{1}{c|}{\multirow{4}{*}{Qwen1.5-14B-Chat}}    & Chat   & 76.99          & 60.19          & 57.81          & 45.31          & 36.32          & 55.32          & 84.49                      & 81.54                               & 82.44          \\
\multicolumn{1}{c|}{}                                     & Plain  & 67.87          & 55.85          & 49.77          & 43.01          & 33.95          & 50.09          & 83.48                      & 79.55                               & 80.76          \\
\multicolumn{1}{c|}{}                                     & FGIV-A & \textbf{79.81} & \textbf{65.28} & \textbf{59}    & 48.94          & 37.06          & \textbf{58.02} & \textbf{86.09}             & 82.12                               & \textbf{83.33} \\
\multicolumn{1}{c|}{}                                     & FGIV-R & 77.42          & 60.37          & 51.99          & \textbf{49.45} & \textbf{42}    & 56.25          & 85.36                      & \textbf{82.31}                      & 83.24          \\ \hline
\multicolumn{1}{c|}{\multirow{4}{*}{Orion-14B-Chat}}      & Chat   & 69.42          & 57.81          & 41.97          & 27.65          & 23.73          & 44.12          & 83.33                      & 73.33                               & 76.4           \\
\multicolumn{1}{c|}{}                                     & Plain  & 66.54          & 55.87          & 45.43          & 45.13          & 31.09          & 48.81          & 82.03                      & 73.46                               & 76.09          \\ 
\multicolumn{1}{c|}{}                                     & FGIV-A & \textbf{73.78} & 60.85          & 52.99          & \textbf{45.18} & 35.29          & 53.62          & 84.35 & \textbf{76.28} & \textbf{78.76} \\
\multicolumn{1}{c|}{}                                     & FGIV-R & 73.44          & \textbf{61.25} & \textbf{53.48} & 43.54          & \textbf{38.27} & \textbf{54}    & \textbf{85.07}             & 75.9                                & 78.71          \\ \hline
\end{tabular}
\end{adjustbox}
\caption{Full experimental results on FollowBench (HSR) and InfoBench.}
\label{tab-all}
\end{table*}
Table~\ref{tab-all} presents the comprehensive experimental results from FollowBench (HSR) and InfoBench. It is evident from these results that models trained with both DPO and SFT generally outperform other baseline models across all base Large Language Models (LLMs).


\section{Prompt Templates}
In this section, we introduce the various prompt templates used with GPT-4 to collect FGIV. Parts of the template formatted as ``\{\{$\cdot$\}\}'' are placeholders that will be replaced with corresponding text chunks, such as instructions or responses. 
\subsection{Filter Invalid Seeds}
\noindent\emph{Please check if the prompt is missing important details that were supposed to be included, or if it is part of a multi-round conversation lacking sufficient context for a standalone response.\\
If either is true, respond with `Yes'. If not, respond with `No'."\\
\\
\text**Given Prompt:**\\
\{\{  Prompt  \}\}\\
\\
\text**Your Judgement (Yes or No):**}

\subsection{Decomposition}
To make GPT-4 better follow the response format, we provide an example inside the prompt template. The decomposed facts and sub-instructions are real response collected from GPT-4.\\

\noindent\emph{Give a prompt, your task is to:\\
1. **Extract Facts:** Identify and list all the factual elements present in the input prompt.\\
2. **Extract Instructions:** List all instructions in the prompt, such as specific actions, constraints, objectives, etc.\\
\\
\text**Given Prompt:**\\
I have to pick up my son. Write a short SMS to my supervisor asking for leaving. In 20 words. Be polite.\\
\\
\text**Extract Facts:**\\
1. The user needs to pick up their son.\\
\\
\text**Extract Instructions:**\\
1. Write a short SMS.\\
2. The SMS is addressed to the user's supervisor.\\
3. The request is for leaving (presumably leaving work or another obligation).\\
4. The message must be polite.\\
5. The message should contain no more than 20 words.\\
\\
\text{**Given Prompt:**}\\
\{\{  Prompt  \}\}}

\subsection{Modification and Reconstruction}
To reduce the cost on data collection, we merge the modification and reconstruction into a single prompt template. We ask GPT-4 to first modify a sub-instruction, then reconstruct the modified sub-instructions into a new instruction variants, and repeat the process for all sub-instructions in the given instruction.\\

\noindent\emph{Below are the given prompt and its extracted facts and instructions.\\
For each instruction in the given extracted instructions perform the following steps:\\
1. **Modify Instruction:** Randomly modify it to a new reasonable one with contradicting semantics or parallel requirements.\\
2. **Revise Prompt:** Only revise the necessary parts of the given prompt to cover the change of instruction. Remain the other parts exactly the same as the given prompt.\\
\\
\text**Given Prompt:**\\
\{\{  Prompt  \}\}\\
\\
\text**Extracted Facts:**\\
\{\{  Extracted Facts  \}\}\\
\\
\text**Extracted Instructions:**\\
\{\{  Extracted Instructions  \}\}\\
\\
\\
Provide an example for each instruction in **Extracted Instructions**. If the modified instruction are unsuitable for an AI language model to respond to, or if the revised prompts are unreasonable or contain invalid instructions, reply `INVALID' for the revised prompt.\\
\text**Modified Instructions 1:**\\
\text{[Modified Instructions]}\\
\\
\text**Revised Prompt 1:**\\
\text{[New Prompts]}\\
\\
\text**Modified Instructions 2:**\\
\text{[Modified Instructions]}\\
\\
\text**Revised Prompt 2:**\\
\text{[New Prompts]}\\
\\
...}

\subsection{Collect Response with Inference}
\noindent\emph{Below is an original prompt and its original response.\\
Now, answer the new prompt by making changes only to the necessary parts of the original response to cover the changes of the new prompt.\\
\\
\text**Original Prompt:**\\
\{\{  Original Prompt  \}\}\\
\\
\text**Original Response:**\\
\{\{  Original Response  \}\}\\
\\
\text**New Prompt:**\\
\{\{  New Prompt  \}\}\\
\\
\text**New Response:**}

\subsection{FGIV-Eval Evaluation}

\noindent\emph{For the provided prompt and its corresponding response below, judge whether the response attempts to address all the instructions included in the prompt.\\
\\
\text**Prompt:**\\
\{\{  prompt  \}\}\\
\\
\text**Response:**\\
\{\{  response  \}\}\\
\\
\text**Your Judgement (answer with 'Yes' or 'No' only):**}

\subsection{FGIV-Det Evaluation}
We employ a specific prompt template to construct FGIV-Det, which is designed to evaluate the LLM's ability to detect instruction non-compliance. Each pair of instruction-response data is inserted into this template, serving as the prompt for LLMs to assess whether the response adheres to the given instruction.\\

\noindent\emph{Evaluate whether the given response adequately addresses all instructions of the given prompt.\\
\\
\text**Given Prompt:**\\
\{\{  Prompt  \}\}\\
\\
\text**Given Response:**\\
\{\{  Response  \}\}\\
\\
\text**Does the response generally meet the instructions of the prompt? Answer with 'Yes' or 'No' only, do not explain:**}

\end{document}